\documentclass{article}

\usepackage{arxiv}

\usepackage[utf8]{inputenc} 
\usepackage[T1]{fontenc}    
\usepackage{hyperref}       
\usepackage{url}            
\usepackage{booktabs}       
\usepackage{amsfonts}       
\usepackage{nicefrac}       
\usepackage{microtype}      
\usepackage{lipsum}		
\usepackage{graphicx}
\usepackage{natbib}
\usepackage{doi}
\usepackage{amsmath,algorithm,tabularx}
\usepackage{algpseudocode}
\usepackage{subcaption}
\usepackage{mwe}
\usepackage{amsmath}
\usepackage[toc,page]{appendix}
\usepackage[labelfont=bf]{caption}
\makeatletter
\newcommand{\multiline}[1]{%
  \begin{tabularx}{\dimexpr\linewidth-\ALG@thistlm}[t]{@{}X@{}}
    #1
  \end{tabularx}
}
\makeatother

\title{ICE-SEARCH: A Language Model-Driven Feature Selection Approach}

\author{ Tianze (Tom) Yang$^*$\\
	School of Computer Science\\
	McGill University, Canada\\
	\texttt{tianze.yang@mail.mcgill.ca} \\
	\And
	Tianyi (Tim) Yang$^*$ \\
	School of Computer Science\\
	McGill University, Canada\\
	\texttt{tianyi.yang@mail.mcgill.ca} \\
 	\AND
	Fuyuan Lyu$^*$ \\
	School of Computer Science\\
	McGill University \& MILA, Canada\\
	\texttt{fuyuan.lyu@mail.mcgill.ca} \\
 	\And
	Shaoshan Liu\dag \\
	Shenzhen Institute of Artificial Intelligence \\
    and Robotics for Society, China\\
	\texttt{shaoshanliu@cuhk.edu.cn}\\
    \And
	Xue (Steve) Liu \\
	School of Computer Science\\
	McGill University, Canada\\
	\texttt{xueliu@cs.mcgill.ca} \\
}

\hypersetup{
pdftitle={ICE-SEARCH},
}

\begin{document}
\def\thefootnote{*}\footnotetext{Equal Contribution}
\def\thefootnote{\dag}\footnotetext{Corresponding Author}
\def\thefootnote{\arabic{footnote}}

\maketitle

\begin{abstract}


This study unveils the \textbf{In-Context Evolutionary Search (ICE-SEARCH)} method, which is among the first works that melds large language models (LLMs) with evolutionary algorithms for feature selection (FS) tasks and demonstrates its effectiveness in Medical Predictive Analytics (MPA) applications. ICE-SEARCH harnesses the crossover and mutation capabilities inherent in LLMs within an evolutionary framework, significantly improving FS through the model's comprehensive world knowledge and its adaptability to a variety of roles. Our evaluation of this methodology spans three crucial MPA tasks: stroke, cardiovascular disease, and diabetes, where ICE-SEARCH outperforms traditional FS methods in pinpointing essential features for medical applications. ICE-SEARCH achieves \textbf{State-of-the-Art (SOTA)} performance in stroke prediction and diabetes prediction; the Decision-Randomized ICE-SEARCH ranks as SOTA in cardiovascular disease prediction. 
The study emphasizes the critical role of incorporating domain-specific insights, illustrating ICE-SEARCH's robustness, generalizability, and convergence. 
This opens avenues for further research into comprehensive and intricate FS landscapes, marking a significant stride in the application of artificial intelligence in medical predictive analytics.

\end{abstract}

\section{Introduction}

The proposed \textbf{In-Context Evolutionary Search (ICE-SEARCH)} method is among the first works that meld large language models (LLMs) with evolutionary algorithms for feature selection (FS) tasks and demonstrates its effectiveness in Medical Predictive Analytics (MPA) applications. 

As a data pre-processing technique, FS provides efficient and effective data preparation for data mining and machine learning problems (\cite{Li_2017}). This procedure can reduce the number of necessary features and facilitate accuracy prediction (\cite{FS_eval_application, massive_data}). For example, FS proves invaluable in stroke prediction, where data is extensive and varied as they are crucial for identifying key predictors among a large number of stroke risk factors (\cite{fs_stroke_1,fs_stroke2,fs_stroke3}). Feature selection (FS) is challenging due to its vast search space, escalating to $2^n$ possible solutions for n features, which exponentially increases computing costs. Common issues with FS techniques include handling large, high-dimensional data inefficiently, limited understanding of data structure, and assumptions about data distribution that may not apply universally. Thus, the evolutionary computation(EC)-based approach (or evolutionary algorithms (EAs)) becomes one of the popular FS algorithms, as it has proven to be an effective solution for feature selection given its global search ability(~\cite{EC4FS_survey}). The EC-based approach treats each FC result as a potential candidate and improves these candidates by manually-designed crossover rules.



The use of Large Language Models (LLMs) has expanded beyond traditional text analysis (\cite{zhao2023survey}), demonstrating versatility and high performance across various sectors, including medical diagnostics (~\cite{medpalm}), financial analysis (~\cite{bloombergGPT}), data preprocessing (~\cite{LLM4DataPrepro}), and code generation (~\cite{codeLLaMA}). This broad applicability highlights LLMs' potential to transform our approach to complex datasets. Specifically, various research has shown the great benefit of using LLM as a cross-over operator in EAs~\cite{chen2023evoprompting}. As previous work (~\cite{guo2023connecting}) suggests, EAs typically modify tokens independently in sequences and ignore the essential connections critical for coherence. In this method, LLMs generate new prompts while EAs oversee the optimization process to select the optimal prompts.  This approach \textbf{reduces the amount of manual design and human bias} involved in the evolutionary algorithm, thereby powering, for instance, evolutionary neuro-architecture search (\cite{chen2023evoprompting}). Moreover, semantic knowledge encoded in the LLM allows for access to a considerable amount of \textbf{cross-domain knowledge} (\cite{ye-etal-2021-crossfit, Sanh2021MultitaskPT, weller-etal-2020-learning}) that would not be available in evolutionary algorithms. More importantly, such knowledge can be enhanced through role-play prompting (\cite{kong2024better}).


Hence, based on the success of EC-based FS approaches and LLM as the cross-over operator, we introduce ICE-SEARCH, an FS methodology that integrates the in-context learning (\cite{Brown2020LanguageMA}) and role-playing capabilities of language models with an evolutionary algorithmic framework. Our method involves engaging the LLM in diverse roles, asking it to identify crucial features for a specific downstream task, and providing the LLM with feedback on the performance of both effective and ineffective feature sets via in-context few-shot prompting, measured in terms of training and validation accuracies of 10-fold cross-validation(CV). Our method improves on traditional feature selection algorithms by merging evolutionary computation with pre-trained knowledge from a large language model. This integration helps us identify important features with improved global search capabilities and validate them without assuming data distribution or requiring a comprehensive understanding of the data structure. To demonstrate the efficacy of our approach, we analyze ICE-SEARCH's performance on the tasks of MPA: stroke prediction, cardiovascular disease prediction, and diabetes prediction- using three increasingly complex datasets.

We summarize our contributions below:

\begin{enumerate}
    \setlength\itemsep{0.05em}
    \item ICE-SEARCH is the first work to apply an LLM to iteratively search for valuable feature sets through role-playing and improve its in-context prompting examples without fine-tuning. 
    \item By integrating an LLM's inherent domain knowledge with an evolutionary algorithm, ICE-SEARCH effectively reduces the search space in MPA datasets and guides the random crossover procedure in an EC-based feature selection approach through its domain knowledge.
    \item ICE-SEARCH achieves SOTA performance in feature selection. Our experiments on three widely adopted MPA tasks demonstrate its effectiveness and robustness. 
\end{enumerate}

\section{Related Work}
\label{sec:related_work}



\textbf{Feature Selection and its Evolutionary Solutions: }
Feature Selection (FS) is a critical dimensionality reduction technique used in data mining and machine learning by selecting a subset of significant features for model construction(~\cite{zebari2020comprehensive}), thereby preserving the original meanings of these features (~\cite{chandrashekar2014survey}). Its primary goal is to mitigate the impact of noise or irrelevant features in datasets, ensuring the accuracy and efficiency of predictions in subsequent models (~\cite{zebari2020comprehensive}). In practice, FS presents itself as a difficult task mainly due to its vast search space, where the total possible solutions are $2^{n}$ for a dataset with n features, leading to exponential growth in computing expenses(~\cite{OptFS}). For example, most of the techniques applied suffer from: (i) huge search space for high-dimensional data whose efficiency deteriorates quickly or computationally infeasible(~\cite{miao2016survey,li2017feature}), (ii) limited knowledge of the underlying data structure(~\cite{li2017feature}), and (iii) imposed case-specific assumption(s) about the probability of the distribution of the data(~\cite{miao2016survey}). To solve (i), evolutionary computation (EC)-based approaches are developed (\cite{FS_survey}) as they are able to effectively search the space of possible feature subsets (\cite{kim2000feature}). Specifically, EC methods for feature selection have been applied in scenarios including \textit{genetic algorithm}(~\cite{holland1992adaptation}), \textit{particle swarm optimization}(~\cite{wang2018particle}), \textit{genetic programming}(~\cite{muni2006genetic}), and \textit{correlation-based filter measures}(~\cite{hall1999correlation}). On the other hand, LLMs have shown great knowledge generalizability (\cite{Wei2021FinetunedLM}) and led to significant research work such as LLM-augmented knowledge graph (\cite{unify_kg_llm}). Thus, our approach surpasses traditional EC methods by combining EC optimization techniques with pre-trained domain knowledge from an LLM. This integration allows us to leverage the pre-trained knowledge to identify significant feature subsets, which we then validate through a downstream task. Unlike other methods, ours does not presuppose any assumptions about the underlying data distribution, effectively addressing issues (i) through (iii).

\textbf{Prompting and Evolutionary Prompting: }
Prompting is a technique employed to give LLMs leverage to address specialized tasks more efficiently. Various state-of-the-art prompting techniques, such as zero-shot or few-shot in-context learning (ICL)(\cite{Brown2020LanguageMA, Min2021MetaICLLT, ICL_Survey,icl_survey2}), role-play prompting, (\cite{xu2023expertprompting, zheng2023a, kong2024better}), Chain-of-Thought(COT) (\cite{Wei2022ChainOT}), Tree-of-Thought(TOT) (\cite{Yao2023TreeOT}), and Graph-of-Thought(GOT) (\cite{Besta2023GraphOT}) have been proven effective in tackling various tasks such as cross-lingual multi-step reasoning \cite{Ranaldi2023EmpoweringMR}.


However, the performance of LLMs varies depending upon the selection of the prompt (~\cite{liu2023pre}), and that leads to automatic prompt optimizations, such as continuous prompt tuning (~\cite{li2021prefix}) and differentiable prompts(~\cite{zhang2021differentiable}). 
In contrast, the high variance in prompt effectiveness for downstream tasks has led to the development of prompt engineering (~\cite{zhou2022large}), prompt revision (~\cite{guo2023learning}), and prompt editing (~\cite{zhang2022tempera}) methods that focus on the exploration or exploitation of candidate prompts to select the most optimal one(s). 

To address the constraints of prompts designed for predefined answers and mitigate the risk of encountering local optima, many have explored evolutionary prompting optimization techniques (~\cite{Yang2023LargeLM, Meyerson2023LanguageMC, guo2023connecting,chen2023evoprompting}). They offer the advantages of not relying on predefined parameters or gradients while simultaneously maintaining a balance between exploration and exploitation.

Such paradigms of evolutionary prompting inspired ICE-Search, which adapts and evolves in the search for optimal prompts via role-playing different experts to identify critical features for downstream tasks. 
Similar to evolutionary prompting, ICE-Search showcases novel ways of leveraging language models via ICL while offering flexibility and adaptability to diverse metrics and tasks.

\textbf{Medical Predictive Analysis: }
Medical Predictive Analytics (MPA) represents a critical and specialized sector within health informatics dedicated to sophisticated medical data examination. MPA leverages advanced statistical algorithms and machine learning methodologies to forecast upcoming trends in healthcare outcomes, including patient admission rates, pharmaceutical advancements, and strategies for disease management (\cite{van2019predictive}). For example, previous works have focused on leveraging various algorithms to facilitate predictive data mining for medical diagnosis(~\cite{soni2011predictive}), disease prediction(~\cite{Usama2019SelfattentionBR}), robotic surgery(~\cite{Oniani2020ArtificialIF}), use of disease diagnosis systems for Electronic Medical Records(~\cite{Latif2020ImplementationAU}), as well as analysis and prediction of outpatient visits(~\cite{Jin2019PredictiveAI}).

\section{Preliminary}
\begin{figure}[t]
    \centering
    \includegraphics[width=0.25\linewidth]{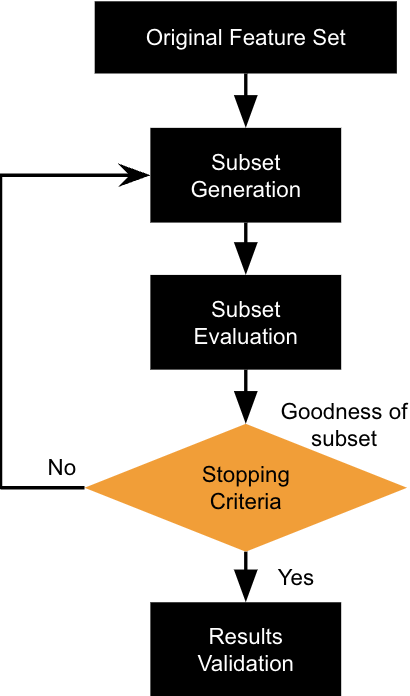}
    \caption{Feature Selection 4-step Process (adapted from \cite{liu2005toward})}
    \vspace{-1pt}
    \label{fig:fs_preliminary}
\end{figure}

In this section, we formulate the feature selection problem. FS is a process of choosing a subset of the original features based on an evaluation criterion that assesses the optimality of each subset. The selection process can be generalized into four steps(~\cite{liu2005toward}), as illustrated in Figure~\ref{fig:fs_preliminary}: subset generation, subset evaluation, stopping criterion, and result validation.

Consider a target task \( T \) and a dataset \( D \) composed of input-output pairs \( (x, y) \in D \) relevant to task \( T \). Define a feature space \( F \) and a probability distribution \( \pi_\theta : F \rightarrow \{0, 1\} \) parameterized by \( \theta \), from which we can sample feature subsets \( S \subseteq F \).  Additionally, we define an evaluation function \( \Phi(S, D): 2^F \times D \rightarrow \mathbb{R} \) that applies a learning algorithm to train a model using the feature subset \( S \) on \( D \), subsequently outputting a real-valued score \( s \). This score \( s \) reflects the effectiveness of \( S \) in terms of accuracy and other performance metrics. The primary objective is to discover a set of feature subsets \( S \sim 2^F \) that, when used to train models on \( D \), optimize the evaluation metric \( \Phi(S, D) \).

The feature selection process includes four key steps:
\begin{enumerate}
    \item \textbf{Subset Generation}: Using the probability distribution \( \pi_\theta \), this step can employ various search strategies to generate candidate feature subsets \( S \subseteq F \). 
    \item \textbf{Subset Evaluation}: Each generated subset \( S \) is evaluated based on a predefined evaluation criterion \( \Phi(S, D): 2^F \times D \rightarrow \mathbb{R} \), which assesses how well the model, trained using features in subset \( S \) on dataset \( D \), performs with respect to the target task \( T \). The criterion can include measures of model accuracy, complexity, etc.
    \item \textbf{Stopping Criterion}:  A stopping criterion indicates a sufficient level of performance and terminates the process of subset generation, and evaluation is repeated. For example, the margin of improvement between iterations, as calculated using a predefined evaluation metric, serves as a stopping criterion.
    \item \textbf{Result Validation}: The best-performing feature subset \( S^* \) is validated and further tested on unseen data or using different validation techniques to ensure the subset's generalizability and prone to overfitting.
\end{enumerate}

Overall, FS identifies an optimal subset \( S^* \) of features from \( F \) that, when used to train a model on \( D \), maximizes the evaluation function \( \Phi(S, D) \) and thus enhances the model's performance on task \( T \).

\section{ICE-SEARCH: Feature Selection with Language Model}

Via zero-shot prompting and in-context learning, ICE-SEARCH aims to streamline the evolutionary training approach for FS in MPA. ICE-SEARCH utilizes an LM as the crossover and mutation operator and directs it to dynamically identify important feature collections through simulating medical practitioners' roles, thus enriching its in-context prompting examples without needing fine-tuning. 


In this section, we explain our algorithm in detail.

\begin{algorithm}
\caption{\textit{In-Context Evolutionary Search} (ICE-SEARCH)}
\footnotesize
\label{pseudocode}
\begin{algorithmic}[1]

\State \textbf{hyperparameters:} $\mathit{Y}$: int, $\mathit{U}$: int, $\mathit{V}$: int, \textit{N}: int, \textit{E}: int
\State \textbf{input:} $\mathcal{D}$, $\mathcal{F}$, $\mathcal{T}$, $\mathcal{Z}$, $\mathcal{X} (\cdot)$, $\mathcal{P}(\cdot)$, $\Psi (\cdot)$, $\Phi (\cdot)$,

\Procedure{}{}
    \State $\mathcal{S} \gets \mathcal{X}(\mathcal{F})$  \label{alg:line:init}
    \State $\mathcal{S}[s] \gets \Phi (\mathcal{D}, s, N)$ for $s$ in $\mathcal{S}$ 
    \State $P \gets \mathcal{P}(\mathcal{T}, \mathcal{F})$ 
    \State $\texttt{TEMPSET}$ = \{\}
    \For{$i \gets 1$ to $\mathit{Y}$}
        \State $s \gets \Psi (P, \text{ROLE})$ 
        \State train acc, val acc = $\Phi (\mathcal{D}, s, N)$ 
        \State $\texttt{TEMPSET}[s] \gets$  train acc, val acc
    \EndFor \label{alg:line:init_end}
    \State $S\text{.append} ({\mathrm{argmax}_s}\  \texttt{TEMPSET}[s][1])$ \label{alg:line:mutation}
    \For{$e \gets 1$ to $\textit{E}$}
     \State P = $\mathcal{P}(\mathcal{T}, \mathcal{F}, S)$ 
        \For{$z$ in $\mathcal{Z}$}
           
            \State \multiline{%
            $f \gets \Psi (P,z)$  }
            \State  train acc, val acc = $\Phi (\mathcal{D}, s, N)$
            \State $\mathcal{S}[s] \gets$ train acc, val acc
        \EndFor \label{alg:line:mutation_end}
        \State Sort $\mathcal{S}$ based on validation accuracy \label{alg:line:filtration}
        \State $\mathcal{S} \gets$  $\mathcal{S}$[:\textit{U}] $\bigcup$ $\mathcal{S}$[-\textit{V}:] \label{alg:line:filtration_end}
    \EndFor
    
    \State 
    $s \gets $ feature set in $\mathcal{S}$ that achieves the highest validation accuracy evaluated by $\Phi$~\label{alg:line:final} 
    \If{ties exist} 
        \State \multiline{%
        choose among the tied feature sets that have the lowest training accuracy}
    \EndIf
    \State return $s$ \label{alg:line:final_end}
\EndProcedure

\end{algorithmic}
\label{alg:ICES_PSEUDO}
\end{algorithm}

\subsection{Algorithm Archetecture}
\begin{figure}[t]
    \centering
    \includegraphics[width=0.7\linewidth]{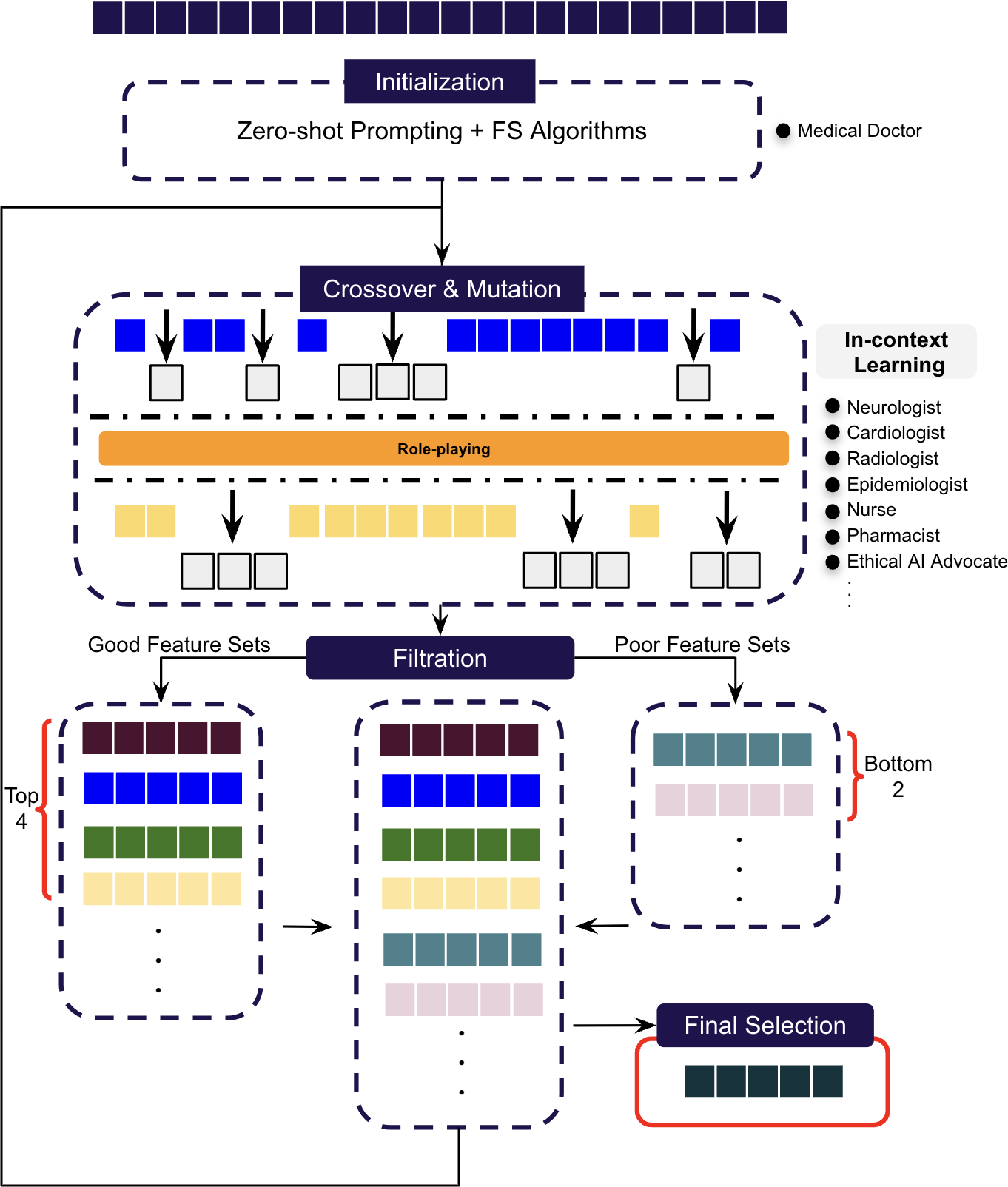}
    \vspace{-3pt}
    \caption{An illustration of ICE-SEARCH Architecture}
    \vspace{-15pt}
    \label{fig:ICE_algo}
\end{figure}

ICE-SEARCH architecture can be broken down into four components: \textit{Initialization}, \textit{Crossover and Mutation}, \textit{Filtration}, \textit{Final Selection}. \textbf{Figure}~\ref{fig:ICE_algo} provides a high-level overview of our algorithm and \textbf{Algorithm}~\ref{alg:ICES_PSEUDO} illustrates its low-level implementation. Notation definitions can be found in Appendix \ref{appendix:notations}

\textbf{Initialization: }
The first stage of ICE-SEARCH is the initialization of a population $\mathcal{S}$, which is a pool of feature sets. This initial population is derived using $\mathit{X}$ number of classical FS methods, with each method's performance assessed based on average training and validation accuracies of the downstream model $\Phi$ over \textit{N-fold} CV. To further refine the initial population, we prompt an LLM $Y$ times in a zero-shot manner, and the output feature combination that yields the highest average validation accuracy over \textit{N-fold} CV is chosen and put in $\mathcal{S}$ for further processing. 
This is discussed in \textit{(line~\ref{alg:line:init}-\ref{alg:line:init_end})}.

\textbf{Crossover and Mutation \textit{(line~\ref{alg:line:mutation}-\ref{alg:line:mutation_end})}:} 
In view of the achievements attributed to \textit{EVO Prompting} by~\cite{chen2023evoprompting} and \textit{EVO Prompt} by \cite{guo2023connecting}, our evolutionary strategy leverages an LLM as the mutation and crossover operator throughout each epoch. Specifically, the LLM assumes $\mathcal{Z}$ distinct roles and is informed about the average training and validation accuracies associated with each feature set in $\mathcal{S}$. During each of these role-plays $\Psi$, the LLM selects a set of features, and its average training and validation accuracies over \textit{N-fold} CV is computed via $\Phi(\cdot)$ and recorded.

\textbf{Filtration \textit{(line~\ref{alg:line:filtration}-\ref{alg:line:filtration_end})}:}
After each epoch, we apply a filtering process. This entails retaining only the feature combinations that fall into two categories: those with the top $\mathit{U}$ highest validation accuracies and those with the bottom $\mathit{V}$ validation accuracies assessed via $\Phi(\cdot)$. This selective mechanism ensures that our algorithm exploits the most promising feature combinations' potential while also maintaining a degree of exploration in $\mathcal{P}$.

\textbf{Final Selection \textit{(line~\ref{alg:line:final}-\ref{alg:line:final_end})}:}
The iterative, evolutionary process is repeated for \textit{E} epochs. Upon completing the final epoch, we select the feature set demonstrating the highest validation accuracy. In cases where multiple feature combinations exhibit the same highest validation accuracy, the tie-breaker is the combination with the lowest training accuracy. This approach ensures that our model performs well on unseen data and avoids overfitting the training dataset. On the other hand, as shown in \textbf{Table}~\ref{tab:ACC_STROKE_SVM_PHI2} and~\ref{tab:ACC_CARDIO_XGB_PHI2}, we have found empirically that all $\mathit{U}$ feature sets are valuable as many of them yield high test accuracy. Therefore, it is not mandatory to follow strictly the strategy we just described. One can customize this component by adjusting steps 24 to 27 in \textbf{Algorithm}~\ref{alg:ICES_PSEUDO} according to the underlying task complexity or dataset quality.

\subsection{Evolution-driven Prompting}
ICE-SEARCH implements a unique and practical evolution-driven prompting mechanism depicted in Figure~\ref{fig:ICE_prompt}. This approach allows the LLM to iteratively generate and evaluate new feature sets, thereby augmenting its own prompting capabilities. This multi-objective functionality offers several key benefits:

\begin{figure}[t]
    \centering
    \includegraphics[width=0.9\linewidth]{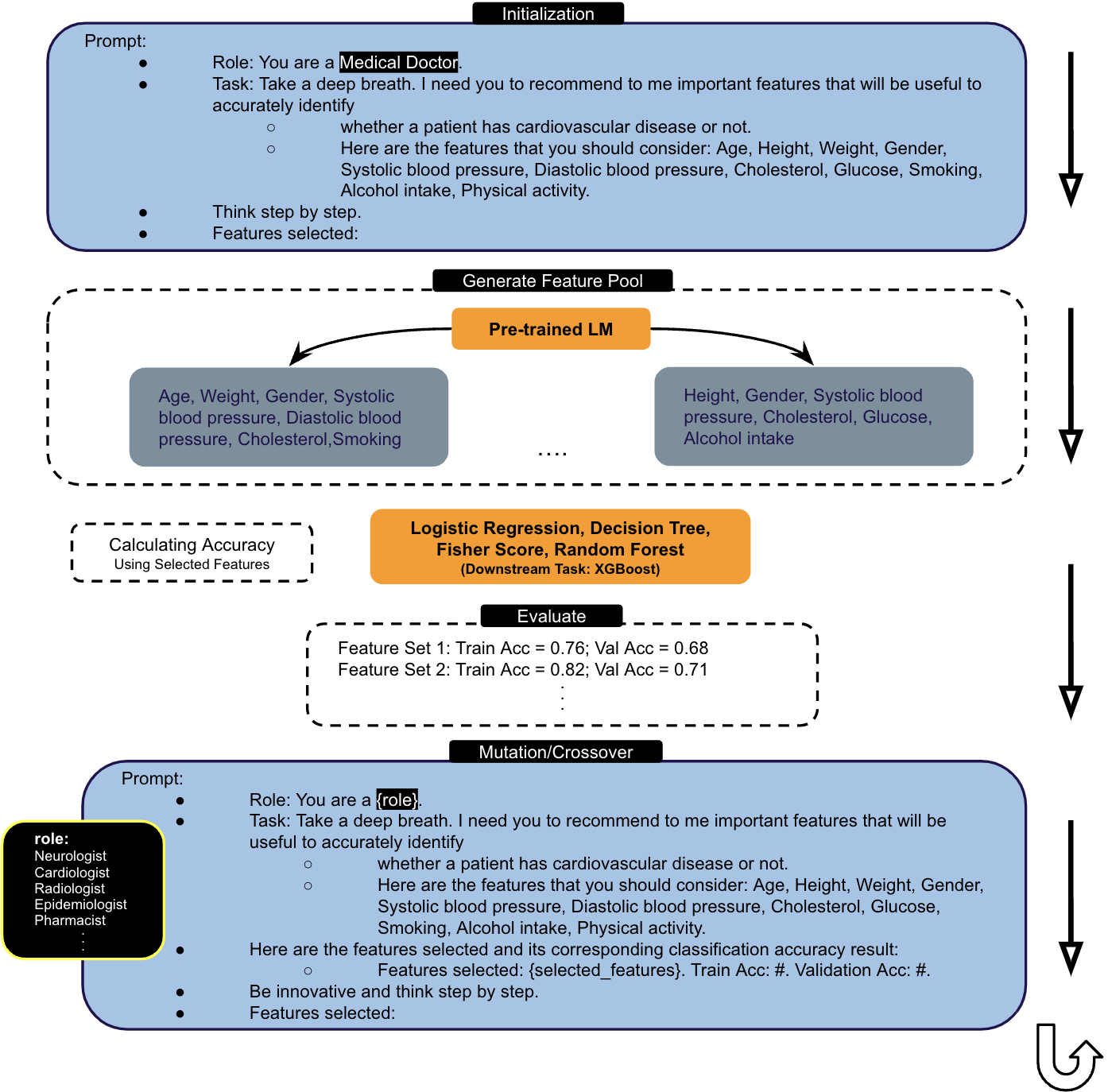}
    \vspace{-5pt}
    \caption{An illustration of ICE-SEARCH Promptings}
    \vspace{-20pt}
    \label{fig:ICE_prompt}
\end{figure}

\textbf{Dynamic Learning and Adaptation:}
By continually generating and assessing new feature sets while eliminating old feature sets, the backbone LLM evolves its understanding of what constitutes an effective feature combination for a given task via in-context learning. This evolutionary learning mechanism ensures that the model remains relevant and creates a continuous improvement loop that augments its prompt over time. Moreover, 

\textbf{Role-played Crossover and Mutation with LLMs}: A suitably prompted and sampled LLM can provide rigorous insights to users' questions, especially through role-playing the characters of facts-based AI assistants(~\cite{shanahan2023role}). This capability stems from models' exposure to factual examples and scientifically precise guidances within its training processes(~\cite{shanahan2023role}). Consequently, the agent excels in role-playing as a reliable source of information, drawing on the factual correctness of statistics-driven and fact-based examples in contexts. Hence, ICE Search leverages an LLM as the crossover and mutation operator by engaging it with various roles. By integrating LLMs directly into the evolutionary search process, we effectively enhance the metrics of interest. This approach allows us to exploit the rich semantic information encoded in LLMs, leading to more informed feature selection decisions. 

\textbf{Customization and Flexibility:}
The adaptability of the algorithm allows for a high degree of customization and flexibility. For example:

\begin{itemize}
    \setlength\itemsep{0.1em}
    \item \textbf{Custom Feature Sets for Different Domains:} In medical datasets, the algorithm can be asked to prioritize features like patient history and laboratory results, whereas in financial datasets, it might focus on market trends and economic indicators.
    \item \textbf{Adaptabililty for Diverse Applications:} The downstream algorithm, population initialization methods, number of epochs, feature pool size, and the roles assumed by the large language model (LLM) can all be tailored based on the task’s complexity and specific requirements. Additional parameters, such as early stopping criteria and maximum/minimum number of features, can also be integrated.
    \item \textbf{Flexible Evolution Strategies:} Depending on the goal, the algorithm can be configured to perform exploitation and exploration based on different policies and metrics. For instance, one might consider including randomly chosen feature sets in the first few epochs.
    \item \textbf{Scalability to Different Dataset Sizes:} Depending on the context length of the backbone LLM, our algorithm can be adjusted to handle large-scale datasets that feature thousands or millions of variables, or it can be fine-tuned for smaller datasets to focus on detailed feature analysis.
\end{itemize}

\subsection{Feature Selection and World Knowledge:}
ICE-SEARCH leverages the LLM's extensive world knowledge and cross-domain expertise to facilitate the transfer of insights across related fields. For example, in the context of cardiovascular disease prediction, the algorithm can draw valuable insights from stroke prediction due to its clinical and biological characteristics. Leveraging knowledge from these related fields enhances the feature selection process, potentially revealing common or complementary predictors that hold significance across various health conditions. Using the vast information stored in an LLM, ICE-SEARCH looks beyond a dataset and helps end-users identify significant features through analysis of interconnected domains.

\section{Experiments and Evaluation}
\label{sec:experiment}

In this section, we demonstrate ICE-SEARCH's effectiveness for FS in three tasks: stroke prediction (~\cite{stroke_dataset}), cardiovascular disease prediction (\cite{cardio_dataset}), and diabetes prediction (~\cite{diabetes_dataset}).



\subsection{Public Health Impact and Significance}

The National Institute of Diabetes and Digestive and Kidney Diseases actively conducts and supports clinical trials targeting diseases with significant public health impact, including hypertensive heart disease (HHD). The global prevalence of HHD and absolute rates of adverse outcomes are expected to continue to rise due to population growth, aging, and continuing increases in obesity globally (\cite{roth2020global}. 

\begin{figure}[t]
    \centering
    \includegraphics[width=0.85\linewidth]{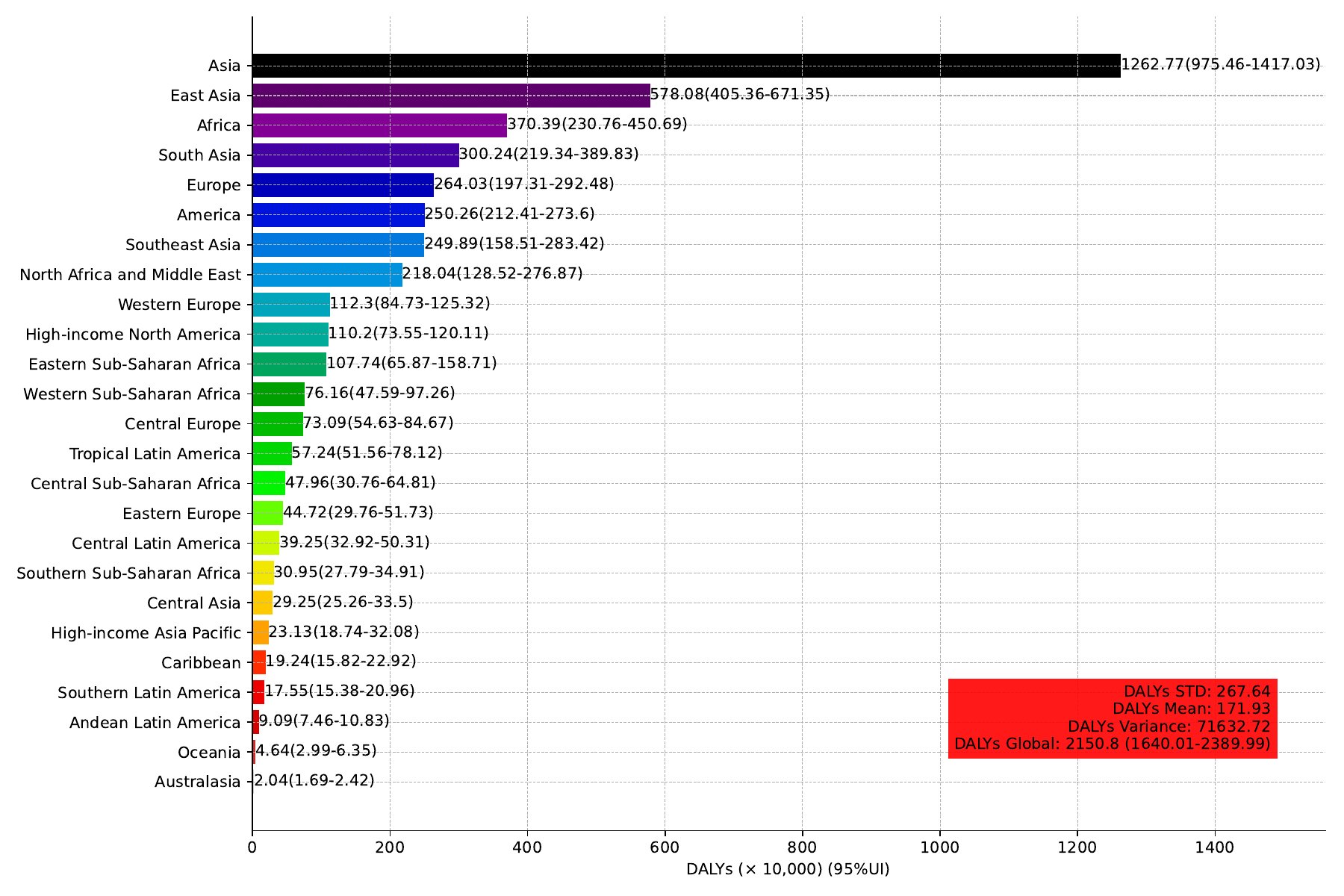}
    \vspace{-10pt}
    \caption{Geometric Patterns of DALYs Due to HHD in 2019. (Source of Data: \cite{yang2023global})}
    \vspace{-10pt}
    \label{fig:hhd}
\end{figure}

Fig~\ref{fig:hhd} demonstrates the heightened global and regional impact of HHD, quantified by the Disability-Adjusted Life Year (DALY) per 100,000 persons — one DALY represents the loss of the equivalent of one year of full health (\cite{WHO_dalys}). The global DALYs attributable to HHD is 2150.8 with a mean of 171.93 and a standard deviation of 267.64 in 2019. They are notably elevated in Asia where DALYs range between 975.46 and 1417.03 (\cite{yang2023global}). 

The observed geographic disparity in DALYs due to HHD highlights the substantial global health burden it represents, particularly in the hardest-hit regions. This disparity underscores the critical need for effective strategies to identify the causes of HHD for both diagnosis and treatment. Current research is exploring a variety of areas, including risk factors specific to certain demographics, advanced diagnostic techniques, and the genetic basis of these conditions. (\cite{NIH}). \textbf{Hence, any improvement on HHD MPA tasks would save a lot of lives.} \textit{As indicated in similar studies, such as~\cite{fekadu2023standard}, high DALYs can be averted as a results of enhanced early treatment likelihood and reduced late treatment-associated mortality. This enhancement helps lower the number of false negatives, thereby reducing the instances of cases that are either missed or diagnosed late. This improvement in diagnosis accuracy can be translated into reduced mortality and DALYs~\cite{fekadu2023standard}. Thus, each percentage improvement in diagnostic accuracy or reduction in errors can translate into significant decreases in DALYs. In fact, if the improved diagnostics lead to a 1\% reduction in DALYs, it would equate to 215,000 life years saved globally based on the 2019 DALYs statistics.}

In this study, we select three open-source datasets on stroke, cardiovascular disease, and diabetes, all of which are either direct complications or contributing factors to HHD, due to the systemic effects of prolonged hypertension on cardiovascular health (\cite{petrie2018diabetes}. These case studies are intended to demonstrate the efficacy of ICE-SEARCH in improving the MPA of diseases with high risk.

\subsection{Datasets}

\textbf{Stroke:} The stroke prediction dataset from Kaggle (~\cite{stroke_dataset}) comprises 10 parameters including patient age, gender, medical history (specifically hypertension and heart disease), marital status, occupation type, residential environment, average glucose level, body mass index, and smoking status. Initially, an imbalanced distribution was observed within the dataset, characterized by a preponderance of negative cases (4861 instances) juxtaposed with positive stroke instances (249 instances). To correct this distributional disparity, we implemented the Synthetic Minority Over-sampling Technique (SMOTE) (~\cite{chawla2002smote}),  which equalizes the representation by creating 4861 cases for stroke and non-stroke instances respectively. Furthermore, to maintain the dataset’s integrity while preserving its inherent distribution, we employed a method of imputation utilizing the median values of the corresponding columns.  

\textbf{Cardiovascular Disease:} The cardiovascular health dataset (~\cite{cardio_dataset}) has a near-equal distribution of both outcomes. It contains 31,783 negative samples and 30,962 positive samples, and includes 11 features including age (in days), height (in centimetres), weight (in kilograms), gender (categorical code), systolic blood pressure, diastolic blood pressure (both in mmHg), cholesterol levels (categorized as normal, above normal, well above normal), glucose levels (similarly categorized), along with binary indicators for smoking, alcohol intake, and physical activity, with no missing values. The primary variable of interest is the presence or absence of cardiovascular disease, represented as a binary target. 

\textbf{Diabetes:} The diabetes dataset (~\cite{diabetes_dataset}) features 218K negative samples, 35K positive samples, and no missing entries. Similarly, to address the class imbalance issue, we applied the SMOTE which results in a balanced dataset with 218,334 negative and 218,334 positive samples. The dataset features 21 indicative variables: high blood pressure, high cholesterol, undergone cholesterol checks, body mass index, smoking status, history of stroke, history of heart disease or attack, engagement in physical activity, regular consumption of fruits, regular consumption of vegetables, heavy alcohol consumption, access to healthcare services, avoiding doctor visits due to cost, general health condition, mental health condition, physical health condition, difficulty in walking, gender, age, education level, income. The target variable has 2 classes, no diabetes and prediabetes/diabetes, represented by 0 and 1 respectively.

\subsection{Experiment Design}
In the experiment, we use Phi2 (~\cite{Javaheripi_Bubeck_2023}) and 4-bit quantized version of LLaMA2 7B(~\cite{touvron2023LLaMA}) models respectively as the backbone LLM, each containing fewer than 8 billion parameters. The rationale for selecting these LLMs lies in their constrained parameter count which allows us to explore the lower boundaries of the performance of ICE-SEARCH.

\textbf{Initialization: }The initialization stage begins by incorporating the backbone LLM and four FS algorithms: Decision Tree, Random Forest, Logistic Regression, and Fisher Score(~\cite{grabczewski2005feature, jaiswal2017application, cheng2006logistic, gu2012generalized}). The LLM is subsequently asked to assume the role of a \textit{medical doctor} to recommend a list of important features based on the given task and dataset features in a zero-shot manner. We prompted the LLM five times, and the best feature set was combined with those selected by the other four FS methods. This combination forms the initial population for our evolution algorithm.

\textbf{Evolution: }The evolution stage comprises 17 distinct roleplays chosen according to the downstream task: \textit{Neurologist, Cardiologist, Radiologist, Epidemiologist, Public Health Professional, Pharmacist, Genetic Counselor, Health Informatics Specialist, Data Scientist, Data Analyst, Machine Learning Engineer, Biostatistician, AI/ML Researcher, Data Engineer, Ethical AI Advocate, Nurse, Emergency Medicine Physician.} The goal of these roleplays is to facilitate crossover and mutation operations within the language models. Note that we did not conduct any prompt optimization or search for the "ideal" roleplays. 

\textbf{Filtration: }During filtration, we use a predetermined downstream prediction model to evaluate average test and validation accuracies over \textit{N-fold} CV of each of the 17 outputs and concatenate them with the feature sets pool. The feature sets pool undergoes an ordering process, whereby only the first five and last three feature sets are retained. At the end of each epoch, the pool contains five good and three poor feature sets. 

\textbf{Downstream Model: }we employ XGBoost (~\cite{chen2016xgboost}) and Support Vector Machine (SVM) as the downstream models and validate the accuracy of the feature sets through a \textit{10-fold} CV to mitigate risks of overfitting. We choose not to conduct an extensive hyperparameter search to emphasize the simplicity and practicality of ICE-SEARCH. The experiments are intended to demonstrate its effectiveness, robustness, generalizability, and potential rather than claiming it as the state-of-the-art methodology for feature selection.

\textbf{Experiment Control:} The experiments are conducted over eight epochs and five seeds(42-46). Key adjustments in the language model parameters include setting the temperature to 1 and the top p-value to 0.9 to introduce controlled randomness into ICE-SEARCH. 

\subsection{Prompt Design}

The algorithm adopts a two-step approach, encompassing zero-shot prompting and few-shot prompting. Note that no optimization is conducted for the prompt design.

\textbf{Zero-Shot Prompting: }During initialization, we present scenarios to the model without any prior training on similar tasks via Zero-Shot prompting. For instance, we use prompts like: \textit{Imagine you are a medical doctor. I need you to recommend important features for accurately [TASK DESCRIPTION]. Consider the following features: [FEATURES]. Think step by step. Selected features: …”.} The model generates responses by relying solely on its pre-existing knowledge.

\textbf{Few-Shot Prompting: }During the crossover and mutation phases of the algorithm, we adopt a more targeted approach via few-shot prompting. We provide samples of feature sets that had achieved the top five and bottom three validation accuracies in previous assessments. An example prompt for this phase is: \textit{As a Neurologist, recommend important features for [TASK DESCRIPTION]. Consider the following features: [FEATURES]. Here are the selected features and their corresponding classification accuracy results: [DETAILED FEATURE AND ACCURACY LIST]. Be innovative and think step by step. Features selected: …}

Both zero-shot and few-shot prompts, as demonstrated in Fig~\ref{fig:ICE_prompt}, are crafted to direct the model's attention to specific tasks and stimulate explorative and rational thinking.

\subsection{Evaluation Results}
\label{sec:experiment_results}

To evaluate the effectiveness of ICE-SEARCH, we rank the testing and average validation accuracies for each possible feature combination in predicting stroke and cardiovascular diseases. Specifically, the stroke dataset includes 1,023 feature combinations, while the cardiovascular disease dataset includes 2,047 combinations. We calculate the validation accuracies using a \textit{10-fold} CV approach and record the average values. However, we do not perform ranking analyses for feature sets in the diabetes dataset. The reason is that assessing test and validation accuracies for each of the $2^{21}$ feature combinations becomes impractical for our study.

\subsubsection{Stroke Prediction:}
\begin{table}[htbp]
\centering
\caption{Average Ranks (out of 1023) of FS Algorithms on \textbf{Stroke} dataset, using \textbf{XGBoost} as Downstream Classifier and \textbf{Phi2} as the Backbone LLM of ICE-SEARCH.}
\resizebox{\textwidth}{!}{%
\begin{tabular}{@{}lccccccc@{}}
\toprule
Metric & Decision Tree & Random Forest & Logistic & Fisher Score & Ensemble of FS & ICE-SEARCH  & Worst ICE-SEARCH\\ \midrule
Test Rank & 149.2 & 48.8 & 77.4 & 93.0 & 51.4 & \textbf{36.4} & 963.8\\
Val Rank & 96.8 & 8.0 & 61.6 & 82.4 & 2.8 & \textbf{1.0} & 977.0\\ \bottomrule
\end{tabular}%
}

\label{tab:RANK_STROKE_XGB_PHI2}
\end{table}

\begin{table}[htbp]
\centering
\caption{Comparative Performance Ranks of ICE-SEARCH on the \textbf{Stroke} Dataset with Different Population Initializations.}
\begin{tabular}{@{}lcc@{}}
\toprule
Population Initialization & Metric & ICE-SEARCH \\ \midrule
FS methods + Phi2 & Test Rank & \textbf{36.4}\\
& Val Rank & \textbf{1.0}\\
Poor Feature Sets + Phi2 & Test Rank & \textit{38.2}\\
& Val Rank & \textit{1.6}\\ \bottomrule
\end{tabular}%

\label{tab:PopIni_ICESEARCH}
\end{table}

\textbf{Table}~\ref{tab:RANK_STROKE_XGB_PHI2} showcases the average rankings for both test and validation accuracies in the context of stroke prediction. These rankings are based on five trials for feature sets chosen by several methods: Decision Tree, Random Forest, Logistic Regression, Fisher Score, and ICE-SEARCH. ICE-SEARCH outperforms the other four feature selection methods with an average test ranking of 36.4 and a validation ranking of 1. This performance not only exceeds that of the individual methods but also surpasses the approach of employing the FS method among the four that produces the feature set with the highest validation accuracy, referred to as the \textit{Ensemble of FS} strategy. 

Interstingly, ICE-SEARCH also showcases its efficacy in identifying less predictive feature sets, with its test and validation accuracies averaging rankings of 963.8 and 977.0, respectively. This highlights ICE-SEARCH's dual capability in distinguishing between effective and ineffective feature sets. This demonstrates its capacity to identify poor feature sets from a baseline of effective ones, thereby implying its effectiveness in uncovering high-quality feature sets even when starting with a low-quality initial population.

To test this hypothesis, we replace the feature sets chosen by the four traditional feature selection methods with four sets with lower accuracy. The training, validation, and testing accuracies of the four feature sets are provided in \textbf{Table}~\ref{tab:bad_start_initialize} in Appendix ~\ref{appendix:poor_init}. As shown in \textbf{Table}~\ref{tab:PopIni_ICESEARCH}, these suboptimal feature sets achieve average validation and test accuracy rankings of 38.2 and 1.6, respectively. The results indicate ICE-SEARCH's ability to target effect feature sets from a baseline of ineffective ones, thus underpinning ICE-SEARCH's robustness against poor initial feature set selection methods.

\begin{table}[!htbp]
\centering
\caption{Test Accuracies(\%) of FS Algorithms and their ranks (out of 1023) on \textbf{Stroke} dataset, using \textbf{XGBoost} as Downstream classifier and \textbf{Phi2} as the Backbone LLM of ICE-SEARCH. ICES $n^{th}$ stands for the feature set with the $n^{th}$ highest validation accuracy chosen by ICE-SEARCH.}
\resizebox{\textwidth}{!}{%
\begin{tabular}{@{}lccccc|ccccc@{}}
\toprule
Seed & Decision Tree & Random Forest & Logistic & Fisher Score & ICES $1^{st}$ & ICES $2^{nd}$ & ICES $3^{rd}$ & ICES $4^{th}$ & ICES $5^{th}$\\ \midrule
42   & 87.969, 23        & 88.072, 21        & 87.506, 65   & 86.221, 333       & \textbf{88.483, 6} & 88.38, 11 & 87.969, 23 & 87.506, 65 & 87.969, 23                            \\ 
43   & 85.758, 418        & \textbf{87.198, 69}        & 86.889, 128  & 86.941, 112       & 86.889, 128 & 86.941, 112 & 86.581, 204 & \textbf{87.558, 27} & 87.095, 84                            \\ 
44   & 88.175, 37       & 88.38, 21         & 88.843, 4   & \textbf{88.997, 1}       & \textbf{88.997, 1} & 88.689, 8 & 88.38, 21 & 87.918, 68 & 88.123, 43                            \\ 
45   & 87.301, 20        & 87.301, 20         & 86.838, 43   & \textbf{87.661, 8}       & \textbf{87.661, 8} & 88.278, 5 & \textbf{88.535, 2} & 87.301, 20 & 87.712, 6                             \\ 
46   & 86.838, 248        & 87.455, 113        & 87.301, 147   & \textbf{88.329, 11}       & 87.969,39 & 87.712, 73 & 87.455, 113 & 87.969, 39 & 87.969, 39                           \\ 
\bottomrule
\end{tabular}%
}

\label{tab:ACC_STROKE_XGB_PHI2}
\end{table}

\textbf{Table}~\ref{tab:ACC_STROKE_XGB_PHI2} presents the test accuracies and their corresponding rankings for four FS methods and ICE-SEARCH. We note that ICE-SEARCH ranks first in three out of five seeds, outperforming/matching the other four FS methods. Remarkably, within the results, ICE-SEARCH's $4^{th}$ and $3^{rd}$ iterations are highlighted as the superior feature sets in seed 43 and 45, respectively. This suggests that not only is ICE-SEARCH $1^{st}$ highly effective, but all top five feature sets identified through the ICE-SEARCH algorithm hold significant value.

\begin{table}[htbp]
\centering
\caption{Average Ranks (out of 1023) of FS Algorithms on \textbf{Stroke} dataset, using \textbf{Support Vector Machine(SVM)} as Downstream Classifier and \textbf{Phi2} as the Backbone LLM of ICE-SEARCH.}
\resizebox{\textwidth}{!}{%
\begin{tabular}{@{}lccccccc@{}}
\toprule
 Metric & Decision Tree & Random Forest & Logistic & Fisher Score & ICE-SEARCH & Worst ICE-SEARCH\\ \midrule
 Test Rank & 58.0 & \textbf{7.8} & 171.2 & 63.4 & 13.8 & 781.2\\
 Val Rank & 38.8 & 7.0 & 201.2 & 52.0 & \textbf{1.4} & 835.8\\
 \bottomrule
\end{tabular}%
}

\label{tab:RANK_STROKE_SVM_PHI2}
\end{table}

\begin{table}[htbp]
\centering
\caption{Test Accuracies(\%) of FS Algorithms and their ranks (out of 1023) on \textbf{Stroke} dataset, using \textbf{SVM} as Downstream classifier and \textbf{Phi2} as the Backbone LLM of ICE-SEARCH. ICES $n^{th}$ stands for the feature set with the $n^{th}$ highest validation accuracy chosen by ICE-SEARCH.}
\resizebox{\textwidth}{!}{%
\begin{tabular}{@{}lccccc|ccccc@{}}
\toprule
Seed & Decision Tree & Random Forest & Logistic & Fisher Score & ICES $1^{st}$ & ICES $2^{nd}$ & ICES $3^{rd}$ & ICES $4^{th}$ & ICES $5^{th}$\\ \midrule
42   & 80.566, 2        & 80.566, 2        & 79.743, 218   & 80.0, 138       & \textbf{80.617, 1} & 80.566, 2 & 80.566, 2 & 80.566, 2 & 80.411, 13                            \\ 
43   & 78.766, 246        & \textbf{79.229, 27}        & 78.766, 246   & 79.126, 63      & 79.126, 63 & 79.126, 63 & \textbf{79.229, 27} & 79.075, 97 & 79.075, 97                            \\ 
44   & 80.36, 33        & \textbf{80.566, 1}         & 80.463, 13   & \textbf{80.566, 1}       & \textbf{80.566, 1} & \textbf{80.566, 1} & \textbf{80.566, 1}   & \textbf{80.566, 1} & 80.514, 7                          \\ 
45   & 79.383, 7        & 79.383, 7         & 79.126, 128   & 79.383, 7       & \textbf{79.434, 2} & 79.383, 7 & 79.383, 7 & 79.383, 7 & 79.383, 7                              \\ 
46   & \textbf{79.846, 2}        & \textbf{79.846, 2}        & 79.229, 251   & 79.537, 108       & \textbf{79.846, 2} & 79.794, 12 & \textbf{79.846, 2} & \textbf{79.846, 2} & \textbf{79.846, 2}                           \\ \bottomrule
\end{tabular}%
}

\label{tab:ACC_STROKE_SVM_PHI2}
\end{table}

\begin{table}[htbp]
\centering
\caption{Average Ranks (out of 1023) of ICE-SEARCH with Different Backbone LLMs on \textbf{Stroke} dataset, using \textbf{XGBoost} as Downstream Classifier.}
\begin{tabular}{@{}lcc@{}}
\toprule
 Metric &  Phi2 & 4bit Quantized LLaMA2 7B \\ \midrule
 Test Rank  & 36.4 & 36.4 \\
 Val Rank  & 1.0 & 1.0 \\
Worst Test Rank   & 963.88 & 1022.6 \\
Worst Val Rank  & 977.0 & 1023.0 \\
\bottomrule
\end{tabular}%

\label{tab:RANK_PHI2_LLaMA2_STROKE_PHI2}
\end{table}

\textbf{Tables}~\ref{tab:RANK_STROKE_SVM_PHI2},~\ref{tab:ACC_STROKE_SVM_PHI2}, and~\ref{tab:RANK_PHI2_LLaMA2_STROKE_PHI2} detail the performance of ICE-SEARCH in stroke prediction, utilizing the Support Vector Machine (SVM) as the downstream classifier and the 4-bit quantized LLaMA 2 7B as the underlying LLM. According to \textbf{Table}~\ref{tab:RANK_STROKE_SVM_PHI2}, although the average test ranking of ICE-SEARCH is 13.8, which is six places lower than the highest average test rank of 7.8, ICE-SEARCH consistently excels in identifying feature sets with the best validation accuracies, achieving an average ranking of 1.4. This performance illustrates ICE-SEARCH's effectiveness not only in selecting effective feature sets but also in recognizing inadequate ones, as evidenced by the significant deterioration in the worst test and validation ranks from 171.2 and 201.2 to 781.2 and 835.8, respectively—an improvement in identifying less effective feature sets by about 110

\textbf{Table}~\ref{tab:ACC_STROKE_SVM_PHI2} further demonstrates ICE-SEARCH's ability to identify the optimal feature sets in four out of five trials. Impressively, it also efficiently discovers additional feature sets that exhibit top test accuracies. Specifically, in seed 42, ICE-SEARCH uncovers one more feature set with the second-highest test accuracy; in seed 46, it finds two additional feature sets also ranking second in test accuracy. This efficiency of ICE-SEARCH in uncovering multiple high-performing feature sets significantly aids researchers in the field of feature selection, offering a broader range of options for achieving optimal test accuracies.

To evaluate the ICE-SEARCH algorithm's adaptability with different underlying Large Language Models (LLMs), its performance was analyzed using a 4-bit quantized LLaMA2 7B. \textbf{Table}~\ref{tab:RANK_PHI2_LLaMA2_STROKE_PHI2} offers a comparative analysis of the ICE-SEARCH feature selection (FS) mechanism when applied to Phi2 and LLaMA 2 7B models. Notably, both models achieved identical rankings for test and validation accuracies. However, the 4-bit quantized LLaMA2 7B demonstrated superior capability in identifying less effective feature sets. Specifically, it accurately identified the feature set with the lowest test and validation accuracies, which were ranked at 1022.6 and 1023, respectively.

\textbf{ICE-SEARCH outperforms conventional feature selection methods in identifying optimal and suboptimal feature sets for stroke prediction. It consistently provides robust and high-quality options across five testing scenarios.}

\subsubsection{Cardiovascular Disease:}
\begin{table}[htbp]
\centering
\caption{Average Ranks (out of 2047) of FS Algorithms on \textbf{Cardiovascular Disease} dataset, using \textbf{XGBoost} as Downstream Classifier and \textbf{Phi2} as the Backbone LLM of ICE-SEARCH.}
\resizebox{\textwidth}{!}{%
\begin{tabular}{@{}lccccccccc@{}}
\toprule
 Metric & Decision Tree & Random Forest & Logistic & Fisher Score & Ensemble of Staitstics & ICE-SEARCH & Worst ICE-SEARCH\\ \midrule
 Test Rank & \textbf{34.4} & 189.8 & 293.2 & 53.0 & 40.0 & 44.6 & 1750.2\\
 Val Rank & 22.6 & 202.0 & 260.4 & 11.4 & 4.2 & \textbf{2.8} & 1761.2\\
 \bottomrule
\end{tabular}%
}

\label{tab:RANK_CARDIO_XGB_PHI2}
\end{table}

\begin{table}[htbp]
\centering
\caption{Test Accuracies(\%) of FS Algorithms and their ranks on \textbf{Cardiovasclar Disease} dataset, using \textbf{XGBoost} as Downstream classifier and \textbf{Phi2} as the Backbone LLM of ICE-SEARCH. ICES $n^{th}$ stands for the feature set with the $n^{th}$ highest validation accuracy chosen by ICE-SEARCH.}
\resizebox{\textwidth}{!}{%
\begin{tabular}{@{}lccccc|ccccc@{}}
\toprule
Seed & Decision Tree & Random Forest & Logistic & Fisher Score & ICES $1^{st}$ & ICES $2^{nd}$ & ICES $3^{rd}$ & ICES $4^{th}$ & ICES $5^{th}$\\ \midrule
42   & \textbf{73.907, 44}        & 73.771, 86        & 73.257, 265   & 73.743, 98       & \textbf{73.907, 44} & 73.743, 98 & \textbf{73.943, 24} & \textbf{73.971, 19} & 73.729, 107                            \\ 
43   & 73.007, 92        & \textbf{73.05, 76}        & 72.586, 316   & 72.95, 120       & 73.007, 92 & 72.95, 120 & \textbf{73.121, 40}  & \textbf{73.079, 64} & 72.921, 137                          \\ 
44   & \textbf{73.279, 1}        & 73.25, 5         & 72.471, 283  & 73.157, 29       & 73.086, 64 & 73.157, 29 & 73.107, 51 & \textbf{73.279, 1} & 73.179, 20                            \\ 
45   & 73.521, 13        & 72.0, 760         & 72.686, 270   & 73.507, 17      & \textbf{73.636, 1} & 73.521, 13 & 73.329, 93  & 73.5, 19 & 73.479, 24                            \\ 
46   & 74.193, 22        & 74.193, 22        & 73.207, 332   & \textbf{74.286, 1}      & 74.193, 22 & 74.057, 85 & 74.2, 18 & 74.207, 13 & \textbf{74.286, 1} \\
\bottomrule
\end{tabular}%
}

\label{tab:ACC_CARDIO_XGB_PHI2}
\end{table}

\textbf{Table}~\ref{tab:RANK_CARDIO_XGB_PHI2} and~\ref{tab:ACC_CARDIO_XGB_PHI2} illustrate the performance of ICE-SEARCH in terms of accuracy and their respective rankings in the cardiovascular disease dataset. In predicting cardiovascular disease, ICE-SEARCH showcases a strong performance with average rankings of 44.6 for test accuracies and 2.8 for validation accuracies. Despite its average testing accuracy ranking second to the four FS methods, ICE-SEARCH ranks first in validation accuracy. This highlights a mismatch between the distributions of the training and testing datasets, suggesting that selecting feature sets based solely on the highest validation accuracy may not guarantee high test accuracy for this dataset. For example, in Table~\ref{tab:ACC_CARDIO_XGB_PHI2}, with the seed set at 42, ICE-SEARCH $3^{rd}$ and $4^{th}$ rank at 24th and 19th, outperforming selections from the Decision Tree and ICE-SEARCH $1^{st}$. Similarly, with seed 43, ICE-SEARCH $3^{rd}$ and $4^{th}$ rank at 40th and 64th respectively in terms of test accuracy, exceeding the performance of selections by Random Forest and ICE-SEARCH $1^{st}$. Furthermore, the lowest test and validation accuracies are ranked at 1750.2 and 1761.2 respectively, emphasizing ICE-SEARCH's consistent ability to identify less significant features.

Given the discrepancy between training, validation and testing distributions, it may be beneficial to adjust steps \textit{~\ref{alg:line:final}-\ref{alg:line:final_end}} in \textbf{Algorithm}~\ref{alg:ICES_PSEUDO} to increase its variance. As a solution, we introduce an adapted version called \textit{Decision-Randomized} ICE-SEARCH, which involves randomly selecting one feature set from the final pool rather than choosing the one with the highest validation accuracy. 

\begin{table}[!htbp]
\centering
\caption{Average Test Ranks (out of 2047) of \textit{Decision-Randomized} ICE-SEARCH on \textbf{Cardiovascular Disease} dataset, using \textbf{XGBoost} as Downstream Classifier and \textbf{Phi2} as the Backbone LLM of ICE-SEARCH. Seven seeds were chosen as a means to maintain the reproducibility of the randomized decisions. ICES $1^{st}$ stands for the feature set with the highest validation accuracy chosen by ICE-SEARCH.}
\begin{tabular}{@{}lccc@{}}
\toprule
Seed &  \textit{Decision-Randomized} ICES & \textit{Decision-Randonmized} \{ICES $\backslash$ ICES $1^{st}$\} \\ \midrule
96   & \textbf{20.0} & 56.2\\
92   & 40.2         & 48.4 \\
88   & 39.4         & 65.2 \\
84   & \textbf{29.4}  & \textbf{31.4}               \\ 
80   & \textbf{29.0}  & 41.6       \\ 
76   & \textbf{24.0}  & 56.2  \\ 
\bottomrule
\end{tabular}%

\label{tab:RANK_CARDIO_RANDOMIZE_ICES}
\end{table}

Table~\ref{tab:RANK_CARDIO_RANDOMIZE_ICES} shows the performance of \textit{Decision-Randomized} ICE-SEARCH. For enhanced reproducibility and reliability, we utilize seven seeds for the random selection process. The findings reveal that in six out of seven cases, \textit{Decision-Randomized} ICE-SEARCH recommends feature sets with an average test accuracy rank lower than those selected by the standard ICE-SEARCH. Notably, in four out of six cases, it identifies feature sets whose performance surpasses those chosen by the four FS methods. This indicates ICE-SEARCH's flexibility to adapt to mismatches between training/validation and testing dataset distributions. 

\textit{Decision-Randomized} ICES, an approach which excludes the first choice of ICE-SEARCH (ICE-SEARCH $1^{st}$) and randomly selects from the subsequent options, demonstrates that in five out of seven instances, it results in less than ideal outcomes. This finding implies that the random selection process is more crucial and that additional layers of selection or filtering may be unnecessary. The experiments underscore the significance of each of the five surviving feature sets, marking the final pool itself as a valuable resource. 

\textbf{The ICE-SEARCH's performance on Cardiovascular Disease dataset achieves an average rankings of 44.6 for test accuracies and 2.8 for validation accuracies on cardiovascular disease datasets. The Decision-Randomized ICE-SEARCH adaptation enhances feature selection across mismatched datasets and optimizes prediction accuracy with a lower average test accuracy rank in six out of seven cases compared to the standard ICE-SEARCH. This adaptation demonstrates ICE-SEARCH’s adaptability and potential for substantial improvement in prediction scenarios.}

\subsubsection{Diabetes Prediction:}
\begin{table}[!htbp]
\centering
\caption{Test Accuracies(\%) of FS Algorithms on \textbf{Diabetes} dataset, using \textbf{XGBoost} as Downstream classifier and \textbf{Phi2} as the Backbone LLM of ICE-SEARCH. ICES $n^{th}$ stands for the feature set with the $n^{th}$ highest validation accuracy chosen by ICE-SEARCH.}
\resizebox{\textwidth}{!}{%
\begin{tabular}{@{}lccccc|ccccccc@{}}
\toprule
Seed & Decision Tree & Random Forest & Logistic & Fisher Score & ICES $1^{st}$ & ICES $2^{nd}$ & ICES $3^{rd}$ & ICES $4^{th}$ & ICES $5^{th}$\\ \midrule
42   & 88.552        & 90.931        & 90.613   & 89.713       & \textbf{90.959} & \textbf{91.002} & 90.927 & 90.931 & 90.897                           \\ 
43   & \textbf{91.017}        & 89.627        & 90.694   & 90.752       & 90.987 & \textbf{91.034}& 90.992 & 91.017 & 90.953                           \\ 
44   & 90.924        & 90.87         & 89.721   & 90.566       & \textbf{90.93} & 90.906 & \textbf{90.934} & 90.87 & 90.924                           \\ 
45   & 88.244        & 90.77         & 90.715   & 90.501       & \textbf{90.899} & 90.77 & 90.81  & 90.77 & 90.747                            \\ 
46   & 90.949        & 90.949        & 90.818   & 90.099       & \textbf{90.995} & 90.949 & 90.818  & 90.93 & 90.936                          \\ 
\bottomrule
\end{tabular}%
}

\label{tab:ACC_DIABETES_XGB_PHI2}
\end{table}

\textbf{Table}~\ref{tab:ACC_DIABETES_XGB_PHI2} presents the comparative performance of FS algorithms on the Diabetes dataset, utilizing XGBoost as a downstream classifier and Phi2 as the LLM. It is important to note that the dataset, after balancing, comprises 436,668 samples and includes 21 indicative variables. This setup enables the examination of ICE-SEARCH's effectiveness in processing and filtering a large volume of information.

In four out of five seeds, ICE-SEARCH $1^{st}$ identifies feature sets with higher test accuracies than those recommended by the four classical FS methods. Furthermore, other feature sets in the final pool of ICE-SEARCH also exhibit high performance. Specifically, at seeds 42 and 43, the ICE-SEARCH $2^{nd}$ discovers feature sets with test accuracies exceeding those selected by the four FS methods and ICE-SEARCH $1^{st}$; at seed 44, ICE-SEARCH $3^{rd}$ identifies the feature set with the highest test accuracy compared to those chosen by the four FS methods and ICE-SEARCH $1^{st}$.

\textbf{ICE-SEARCH consistently demonstrates superior performance in feature selection and surpasses traditional methods in multiple instances. It consistently achieves higher test accuracies than conventional feature selection methods across various trials, effectively identifying optimal feature sets that provide the most accurate diabetes diagnosis at numerous seeds.}

In summary, ICE-SEARCH consistently outperforms traditional feature selection (FS) methods and is capable of identifying a greater number of high-quality feature sets from a large dataset. These findings provide empirical evidence that establishes the ICE-SEARCH as a superior FS technique compared to conventional FS methods.

\subsection{Rate of Convergence:}

ICE-SEARCH algorithm demonstrates fast convergence characteristics in isolating optimal sets of features. A detailed analysis can be found in Appedix \ref{appendix:converge}

Overall, the convergence is approximately attained by Epoch three, five, and one in the three cases. These convergence patterns observed are indicative of the swift learning capabilities of ICE-SEARCH. Specifically, the initial steep slope observed in each chart, followed by a plateau phase, highlights the efficiency of early learning. It is also worth noting that such patterns of convergence are desirable as they indicate a model's ability to rapidly approximate the optimal function without the need for excessive training.

\section{Discussion} 

To effectively improve the metrics of interest and search for valuable feature sets, ICE-SEARCH employs an LLM for crossover and mutation operations and classical FS algorithms for initialization. We have noticed five key advantages of our proposed approach:

\textbf{High generalizability:} ICE-SEARCH can be readily adapted to a variety of tasks and algorithms. As evidenced by \textbf{Tables} (\ref{tab:RANK_STROKE_XGB_PHI2}, \ref{tab:RANK_STROKE_SVM_PHI2}, \ref{tab:RANK_PHI2_LLaMA2_STROKE_PHI2}), ICE-SEARCH continues to output high-performance feature subsets when integrated with different backbone LLMs and downstream predictors, shedding light on the generalizability of ICE-SEARCH. Additionally, \textbf{Table} \ref{tab:ACC_DIABETES_XGB_PHI2} showscases our algorithm's ability to manage high-dimensional and high-volume datasets.

\textbf{Robust against Poor Initialization: }As illustrated in \textbf{Tables} (\ref{tab:RANK_STROKE_XGB_PHI2}, \ref{tab:bad_start_initialize}), ICE-SEARCH has shown robustness to ineffective initialization methods and maintained its feature selection efficiency across three distinct downstream binary classification tasks. It exhibits the capability to adapt to various initialization methods while preserving robustness against the quality of the initial population. 

\textbf{Adaptability:} ICE-SEARCH can be easily adapted to datasets with imbalanced training and testing distributions. For example, \textbf{Table} \ref{tab:RANK_CARDIO_RANDOMIZE_ICES} shows that its alternative, \textit{Decision-Randomized} ICE-SEARCH maintains high performances despite differences in the distributions of training and test datasets, particularly evident in the cardiovascular disease dataset. This demonstrates textit{Decision-Randomized} ICE-SEARCH's robustness to discrepancies between the distributions of training and test datasets, thereby highlighting the overall adaptability of ICE-SEARCH. 

\textbf{Fast Convergence}: As shown in Fig \ref{fig:all_convergence}, on all three datasets, ICE-SEARCH achieves convergence within three to four epochs, showcasing an efficient convergence rate in identifying the optimal feature sets. 

\textbf{Light-weight training design}:
ICE-SEARCH leverages the extensive world knowledge pre-trained into an LLM to iteratively refine a population of feature sets. Figs \ref{fig:ICE_algo} and \ref{fig:ICE_prompt} provide a high-level illustration of our implementtaion of ICE-SEARCH. It employs strategies such as role-playing and few-shot learning without the need for actual fine-tuning. In addition, more advanced techniques such as Tree-of-Thoughts prompting can be applied to the framework (~\cite{T_o_T}). This training approach effectively reduces the computational demands associated with training high-performance deep learning (DL) models, as highlighted by ~\cite{thompson2020computational}.

\section{Conclusion}

LLMs with their adeptness at encapsulating vast amounts of world knowledge and facilitating cross-domain insights, have demonstrated exceptional utility across a wide range of fields. This paper introduces ICE-SEARCH, which for the first time explores the integration of LLMs with an evolutionary search framework for FS in MPA tasks. 

ICE-SEARCH harnesses the crossover and mutation capabilities inherent in LLMs within an evolutionary framework, significantly improving FS through the model's comprehensive world knowledge and its adaptability to a variety of roles. Our evaluation of this methodology spans three crucial MPA tasks: stroke, cardiovascular disease, and diabetes, where ICE-SEARCH outperforms traditional FS methods in pinpointing essential features for medical applications. 




\bibliographystyle{unsrtnat}
\bibliography{main_arxiv}  
\newpage
\appendix
    
\section{Appendix}
\subsection{Notations of ICE-Search}
\label{appendix:notations}
\begin{table}[htbp]
\centering
\caption{Summary of Notations used in \textbf{Algorithm} \ref{alg:ICES_PSEUDO}}
\resizebox{\textwidth}{!}{%
\begin{tabular}{c l}
\hline
\textbf{Notation} & \textbf{Description} \\
\hline
$\mathcal{X}$ & A mapping that uses a set of $X$ number of classical FS methods used to initialize the feature set pool. Output $X$ feature sets. \\
$\textit{Y}$ & Number of times for the LLM to select feature sets through zero-shot prompting for feature set pool initialization. \\
$\mathcal{Z}$ & Set of role-plays for the LLM intended for evolution. \\
$\textit{U}$ & Number of top feature sets to keep. \\
$\textit{V}$ & Number of bottom feature sets to keep. \\
$N$ & Number of groups that the given dataset is split into for CV. \\
$E$ & Number of Epochs. \\
$\mathcal{F}$ & The given feature set \\
$\mathcal{S}$ & Feature set pool (i.e. population) \\
$\mathcal{D}$ & Dataset. \\
$\mathcal{T}$ & Task description. \\

\hline
\end{tabular}
}

\label{table:notations}
\end{table}

\begin{table}[htbp]
\centering
\caption{Summary of Mapping Functions used in \textbf{Algorithm} \ref{alg:ICES_PSEUDO}}
\resizebox{\textwidth}{!}{%
\scriptsize %
\begin{tabular}{p{0.12\linewidth} | p{0.6\linewidth}}
\hline
\textbf{Function} & \textbf{Description} \\
\hline
$\Phi(\mathcal{D}, s, N)$ & Evaluation of the input feature set $f$ using a pre-determined downstream model on dataset $\mathcal{D}$, output average training and testing accuracies across $N$-fold CV. \\
$\Psi(P, z\text{=None})$ & A mapping that prompts a Language model to output a feature set. Users can choose which role they want the LLM to assume. \\
$\mathcal{P}(\mathcal{T}, \mathcal{F}, \mathcal{S}\text{=None})$ & A mapping from task description, original feature set, sample feature subset (optional) to a prompt $P$ used to instruct an LLM to perform feature selection. \\
\hline
\end{tabular}
}

\label{table:notations_cont}
\end{table}

\newpage
\subsection{Poor Population Initialization}
\label{appendix:poor_init}

\begin{table}[!htbp]
\centering
\caption{The accuracy of the four suboptimal feature sets was selected to evaluate the robustness of ICE-SEARCH against poor feature set initialization.}
\begin{tabular}{@{}lccccc|ccccc@{}}
\toprule
Seed & Train & Valid & Test \\
\midrule
42 & 71.174 & 69.436 & 68.792 \\
 & 68.314 & 65.475 & 65.810 \\
 & 81.358 & 79.349 & 81.028 \\
 & 64.758 & 63.893 & 63.753 \\
\midrule
43 & 70.966 & 68.908 & 69.717 \\
 & 68.353 & 65.976 & 66.324 \\
 & 81.910 & 80.017 & 79.640 \\
 & 64.579 & 63.945 & 64.576 \\
\midrule
44 & 71.271 & 69.422 & 67.969 \\
 & 68.243 & 65.334 & 66.787 \\
 & 81.658 & 79.786 & 80.154 \\
 & 64.675 & 64.241 & 64.216 \\
\midrule
45 & 71.053 & 69.436 & 69.409 \\
 & 68.363 & 65.848 & 65.964 \\
 & 81.498 & 79.606 & 80.668 \\
 & 64.644 & 63.958 & 64.165 \\
\midrule
46 & 70.871 & 68.638 & 70.026 \\
 & 68.141 & 65.436 & 66.530 \\
 & 81.525 & 79.530 & 80.771 \\
 & 64.279 & 64.279 & 64.833 \\
\bottomrule
\end{tabular}%
\label{tab:bad_start_initialize}
\end{table}
\subsection{Rate of Convergence}
\label{appendix:converge}

\begin{figure}[hbt!]
    \centering
    \begin{subfigure}[b]{0.6\textwidth}
        \centering
        \includegraphics[width=\linewidth]{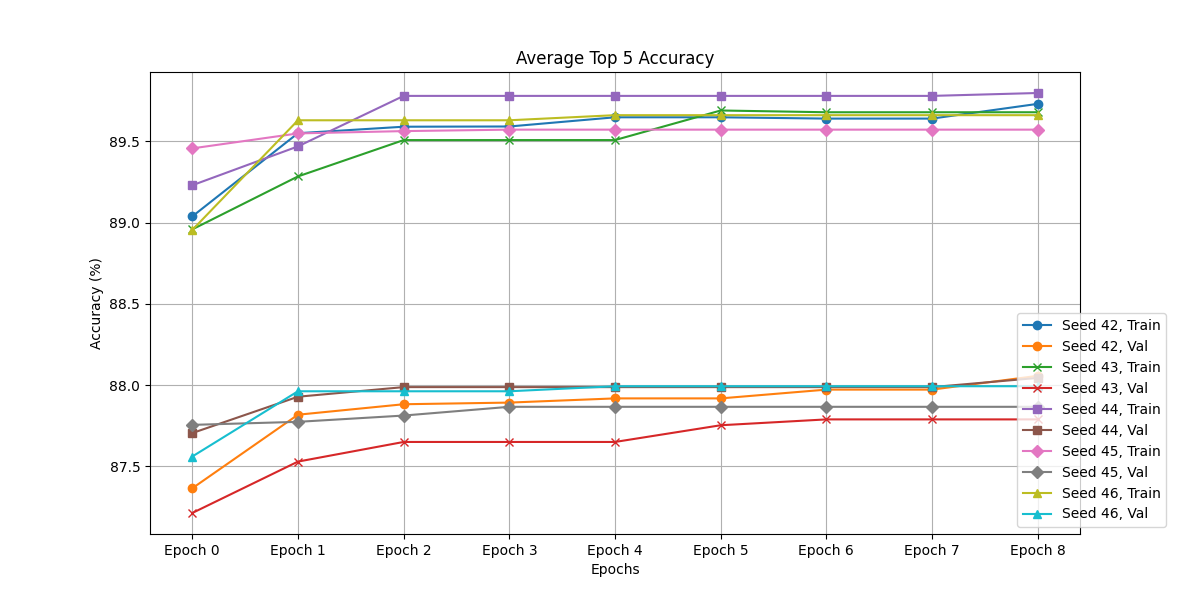}
        \caption{Average Top 5 Training and Validation Accuracies from 10-Fold Cross-Validation for the Stroke Dataset}
        \label{fig:stroke_convergence}
    \end{subfigure}
    \hfill 
    \begin{subfigure}[b]{0.6\textwidth}
        \centering
        \includegraphics[width=\linewidth]{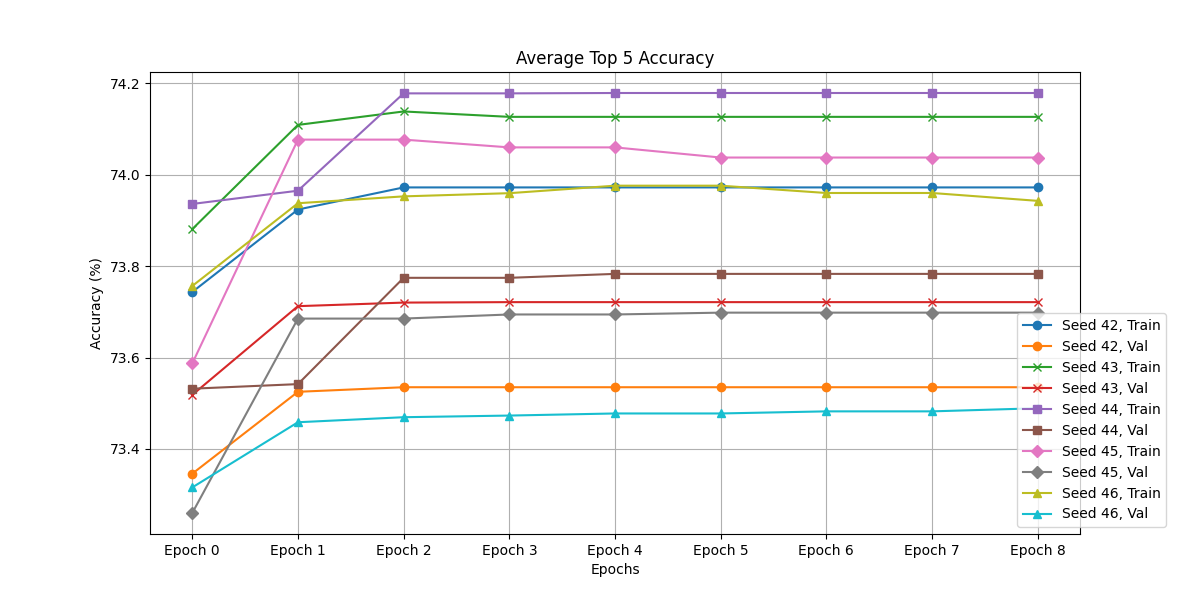}
        \caption{Average Top 5 Training and Validation Accuracies from 10-Fold Cross-Validation for the Cardiovascular Disease Dataset}
        \label{fig:cardio_convergence}
    \end{subfigure}
    \begin{subfigure}[b]{0.6\textwidth} 
        \centering
        \includegraphics[width=\linewidth]{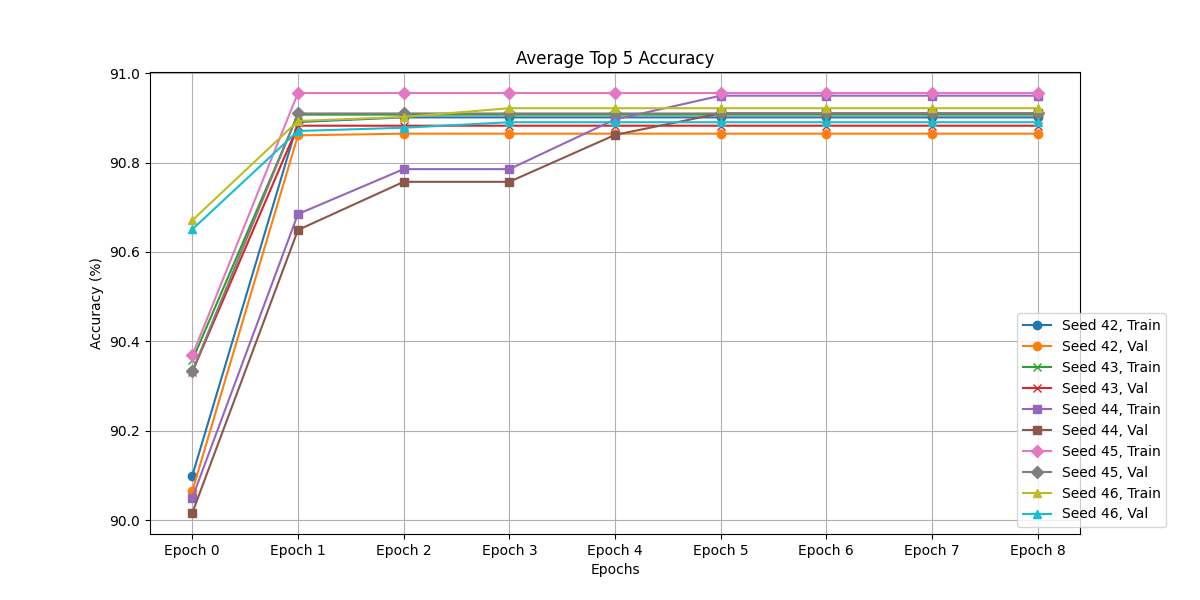}
        \caption{Average Top 5 Training and Validation Accuracies from 10-Fold Cross-Validation for the Diabetes Dataset}
        \label{fig:diabetes_convergence}
    \end{subfigure}
    \caption{Comparative convergence analysis of XGBoost model accuracies across stroke, cardiovascular, and diabetes datasets over 10-fold cross-validation. The top row displays the convergence trends for the stroke dataset, the middle for the cardiovascular dataset, and the bottom figure for the diabetes dataset. This figure highlights the model's performance and convergence behavior in handling different medical datasets through cross-validation.}
\label{fig:all_convergence}
\end{figure}
As shown in Fig~\ref{fig:stroke_convergence}, the accuracy increases steadily until approximately Epoch three, after which the rate of improvement diminishes and the lines become parallel to the epoch axis, suggesting that the model is nearing their performance limits.

In Fig~\ref{fig:cardio_convergence}, the accuracy improves consistently over a larger number of epochs, not plateauing until around Epoch five. This suggests a slower learning process, possibly due to a more complex feature structure requiring more iterations to reach optimal performance.


In Fig~\ref{fig:diabetes_convergence}, it demonstrates a rapid initial improvement in accuracy, with convergence mostly occurring by Epoch one. Post Epoch one, the accuracy levels off, indicating that the model located a solution space that did not significantly improve with further training.

In each case, convergence is indicated by the levelling out of accuracy improvements, where additional epochs do not result in significant gains. 

\end{document}